\documentclass[letterpaper, 10 pt, conference]{ieeeconf}  % Comment this line out
\IEEEoverridecommandlockouts                              % This command is only
\usepackage[utf8]{inputenc}
\usepackage{gensymb}
\usepackage{booktabs}
\newcommand{\ra}[1]{\renewcommand{\arraystretch}{#1}}
\usepackage[acronym]{glossaries}
\newacronym{pid}{PID}{proportional-integral-derivative}
\newacronym[prefixfirst={a\ },% prefix used on first use
            prefix={an\ }% prefix used on subsequent use
            ]{ml}{ML}{machine learning}
\newacronym{drl}{DRL}{deep reinforcement learning}
\newacronym{dnn}{DNN}{deep neural network}
\newacronym{mlp}{MLP}{multi layer perceptron}
\newacronym[prefixfirst={a\ },% prefix used on first use
            prefix={an\ }% prefix used on subsequent use
            ]{nn}{NN}{neural network}
\newacronym[prefixfirst={a\ },% prefix used on first use
            prefix={an\ }% prefix used on subsequent use
            ]{rnn}{RNN}{recurrent neural network}
\newacronym[prefixfirst={a\ },% prefix used on first use
            prefix={an\ }% prefix used on subsequent use
            ]{rl}{RL}{reinforcement learning}
\newacronym{ppo}{PPO}{proximal policy optimization}
\newacronym{ddpg}{DDPG}{deep deterministic policy gradient}
\newacronym{trpo}{TRPO}{trust region policy optimization}
\newacronym{sac}{SAC}{soft actor-critic}
\newacronym{dqn}{DQN}{deep Q-network}
\newacronym{mdp}{MDP}{Markov decision process}
\newacronym{uav}{UAV}{unmanned aerial vehicle}
\newacronym{auv}{AUV}{autonomous underwater vehicle}
\newacronym{fvi}{FVI}{fitted value iteration}
\newacronym{nmpc}{NMPC}{nonlinear model predictive control}
\newacronym{mrac}{MRAC}{model reference adaptive control}
\newacronym{ndi}{NDI}{nonlinear dynamic inversion}
\newacronym{smc}{SMC}{sliding mode control}
\newacronym{dof}{DOF}{degree of freedom}
\newacronym[prefixfirst={a\ },% prefix used on first use
            prefix={an\ }% prefix used on subsequent use
            ]{mpc}{MPC}{model predictive controller}
\newacronym{kl}{KL}{Kullback–Leibler}
\newacronym{lstm}{LSTM}{long short term memory}
\newacronym[prefixfirst={a\ },% prefix used on first use
            prefix={an\ }% prefix used on subsequent use
            ]{mc}{MC}{Monte Carlo}
\newacronym{gae}{GAE}{generalized advantage estimate}
\newacronym{td}{TD}{temporal difference}

\usepackage{graphicx}
\usepackage{subcaption} % TODO: might have to use subfig instead since apparently some latex templates are incompatible with subcaption
\usepackage{multirow}
\usepackage{natbib}
\usepackage{todonotes}

\usepackage{xcolor}
\newcommand\myworries[1]{\textcolor{red}{#1}}
\usepackage{amsmath} % assumes amsmath package installed

\title{\LARGE \bf
Attitude Control of Fixed Wing UAVs with Reinforcement Learning
}

\author{Eivind Bøhn$^{1}$, Erlend Coates, Signe Moe, Tor Arne Johansen$^{2}$% <-this % stops a space
\thanks{*This work was not supported by any organization}% <-this % stops a space
\thanks{$^{1}$H. Kwakernaak is with Faculty of Electrical Engineering, Mathematics and Computer Science,
        University of Twente, 7500 AE Enschede, The Netherlands
        {\tt\small h.kwakernaak at papercept.net}}%
\thanks{$^{2}$P. Misra is with the Department of Electrical Engineering, Wright State University,
        Dayton, OH 45435, USA
        {\tt\small p.misra at ieee.org}}%
}

\begin{document}

\maketitle
\thispagestyle{empty}
\pagestyle{empty}

%%%%%%%%%%%%%%%%%%%%%%%%%%%%%%%%%%%%%%%%%%%%%%%%%%%%%%%%%%%%%%%%%%%%%%%%%%%%%%%%
\begin{abstract}
Contemporary autopilot systems for \glspl{uav} are far more limited in their flight envelope as compared to experienced human pilots, thereby restricting the conditions \glspl{uav} can operate in and the types of missions they can accomplish autonomously. This paper proposes a \gls{drl} controller to handle the nonlinear attitude control problem, enabling extended flight envelopes for fixed-wing \glspl{uav}. A proof-of-concept controller using the \gls{ppo} algorithm is developed, and is shown to be capable of stabilizing a fixed-wing \gls{uav} from a large set of initial conditions to reference roll, pitch and airspeed values. The training process is outlined and key factors for its progression rate are considered, with the most important factor found to be limiting the number of variables in the observation vector, and including values for several previous time steps for these variables. The trained \gls{rl} controller is compared to a \gls{pid} controller, and is found to converge in more cases than the \gls{pid} controller, with comparable performance. Furthermore, the \gls{rl} controller is shown to generalize well to unseen disturbances in the form of wind and turbulence, even in severe disturbance conditions.
\end{abstract}

% TODO: til tilbakemeldere:
% Reading guide (my style is just writing, so please give input on what can be removed or relocated
% Ask about the use of abbrevations (i use glossary, but sometimes it feels unnatural to just have the thing spelled out once)
% Ask for input on disposition
% Point at that since it is a first draft, not much effort has been put into making it pretty (figures/tables etc.), but if they have some input on the content of figures/tables/captions etc. that is much appreciated. 

\listoftodos
%%%%%%%%%%%%%%%%%%%%%%%%%%%%%%%%%%%%%%%%%%%%%%%%%%%%%%%%%%%%%%%%%%%%%%%%%%%%%%%%
\section{INTRODUCTION}\label{sec:introduction} 
\Glspl{uav} are employed extensively to increase safety and efficiency in a plethora of tasks such as infrastructure inspection, forest monitoring, and search and rescue missions. Many tasks can however not be accomplished fully autonomously, due to several limitations of autopilot systems. Low-level stabilization of the \gls{uav}'s attitude provided by the inner control loops is increasingly difficult, due to various nonlinearities, as the attitude and airspeed deviates from stable, level conditions. The outer control layers providing path planning and guidance has to account for this, and settle for non-agile and safe plans. Equipping the autopilot with the stabilization skills of an experienced pilot would allow fully autonomous operation in turbulent or otherwise troublesome environments, such as search and rescue missions in extreme weather conditions, as well as increasing the usefulness of the \gls{uav} by for instance allowing the \gls{uav} to fly closer to its targets for inspection purposes.

\begin{figure}[thpb]
        \includegraphics[width=0.48\textwidth]{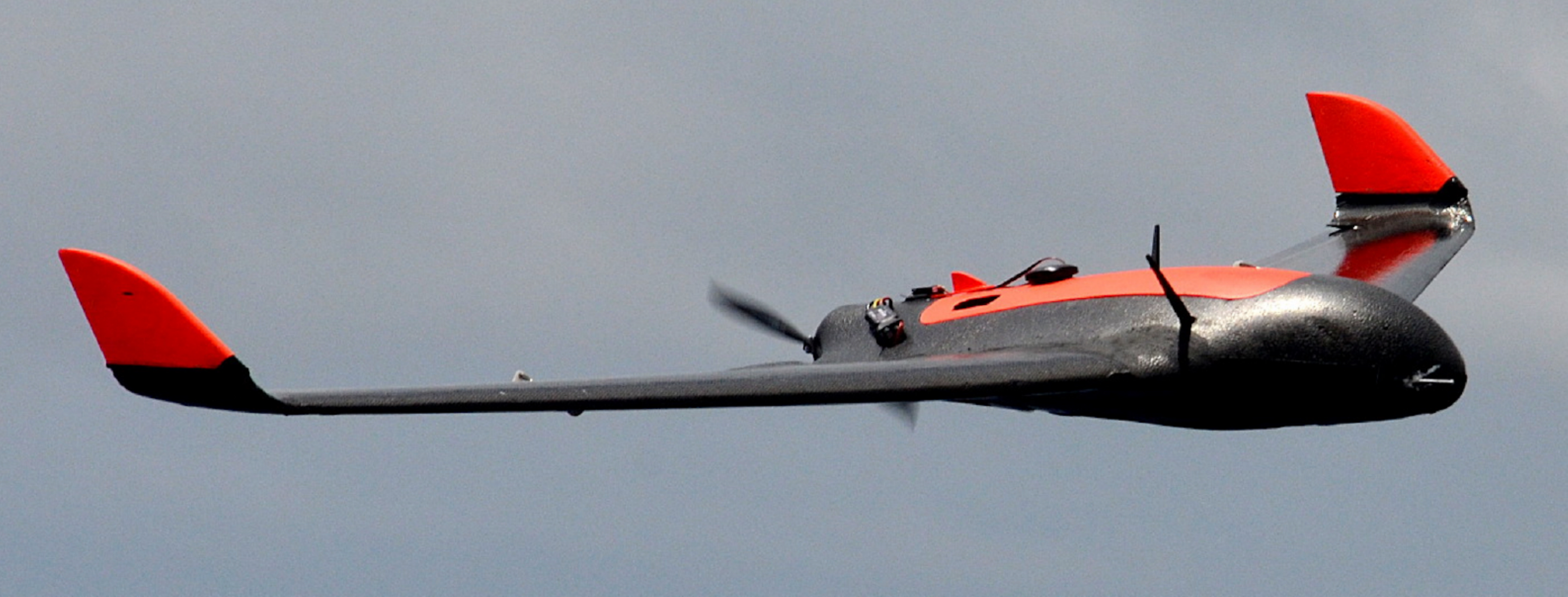}
        \caption{Skywalker X8 Fixed-Wing UAV}
        \label{fig:x8}
\end{figure}

%The workhorse behind the inner loop of conventional autopilots, the ubiquitous \gls{pid} controller, is linear in its nature and therefore requires the highly non-linear aerodynamics affecting the aircraft during flight to be linearized. This idealization introduces artifacts which are major contributing factors in the described discrepancy between a human pilot and the autopilot. The accuracy of the linearized model quickly degrades when diverging from the linearization point, necessitating several linearized models and accompanying \gls{pid} controller gains for different operating circumstances, with subsequent switching effects.
Autopilots for fixed-wing \glspl{uav}, as illustrated in Figure~\ref{fig:x8}, are typically designed using cascaded single-variable loops under assumptions of decoupled longitudinal and lateral motion, using classical linear control theory~\cite{Beard}. The dynamics of fixed-wing aircraft including \glspl{uav} are however strongly coupled and nonlinear. Nonlinear terms in the equations of motion include kinematic nonlinearities (rotations and coriolis effects), actuator saturation and aerodynamic nonlinearities, which are also uncertain and difficult to model. The decoupled and linear designs are reliable and well-tested for nominal flight, but also requires conservative safety limits in the allowable range of flight conditions and maneuvers (flight envelope protection), because linear controllers applied to nonlinear systems typically result in a limited region of attraction~\cite{Khalil2001}. This motivates the use of state-of-the-art nonlinear control algorithms.

Examples of nonlinear control methods applied to \glspl{uav} include gain scheduling~\cite{Girish2014}, linear parameter varying (LPV) control~\cite{Rotondo2017}, dynamic inversion (feedback linearization)~\cite{Kawakami2017}, adaptive backstepping~\cite{Ren2005}, sliding mode control~\cite{Castaneda2017}, nonlinear model predictive control~\cite{Mathisen2016}, nonlinear H-infinity control~\cite{Garcia2017}, dynamic inversion combined with mu-synthesis~\cite{Michailidis2017}, model reference adaptive control~\cite{EugeneLavretsky2012} and L1 adaptive control~\cite{kaminer}. Automated agile and aerobatic maneuvering is treated in~\cite{Levin2019} and~\cite{Bulka2019}. Several of these methods require a more or less accurate aerodynamic model of the UAV. A model-free method based on fuzzy logic can be found in~\cite{Kurnaz2008}. Fuzzy control falls under the category of intelligent control systems, which also includes the use of neural networks. An adaptive backstepping controller using a neural network to compensate for aerodynamic uncertainties is given in~\cite{Lee2001}, while a genetic neuro-fuzzy approach for attitude control is taken in~\cite{Oliveira2017}. The state of the art in intelligent flight control of small \glspl{uav} is discussed in~\cite{Santoso2018}.

%\gls{ml} methods therefore seem like a natural candidate to be involved in some capacity of the improved autopilot, whether it be in a system identification role or directly in the control aspect.
Control of small \glspl{uav} requires making very fast control decisions with limited computational power available. Sufficiently sophisticated models incorporating aerodynamic nonlinearities and uncertainties with the necessary accuracy to enable robust real-time control may not be viable under these constraints. Biology suggests that a bottom-up approach to control design might be a more feasible option. Birds perform elegant and marvelous maneuvers and are able to land abruptly with pinpoint accuracy utilizing stall effects. Insects can hover and zip around with astonishing efficiency, in part due to exploiting unsteady, turbulent aerodynamic flow effects~\cite{Beard}. These creatures have developed the skills not through careful consideration and modeling, but through an evolutionary trial-and-error process driven by randomness, with mother nature as a ruthless arbiter of control design proficiency. In similar bottom-up fashion, \gls{ml} methods have shown great promise in uncovering intricate models from data and representing complex nonlinear relations from its inputs to its outputs. \gls{ml} can offer an additional class of designs through learning that are not easily accessible through first principles modeling, exhibiting antifragile properties where unexpected events and stressors provide data to learn and improve from, instead of invalidating the design. It can harbor powerful predictive powers allowing proactive behaviour, while meeting the strict computation time budget in fast control systems.
%Control of a small \gls{uav} requires fast actuation decisions many times a second, but simplified models suitable for real-time control --- incorporating turbulence and the delayed effects of actuation while being valid for the strongly nonlinear aerodynamics --- may not be viable. Still, biology shows us that mastering and even exploiting these phenomena is possible: 

%While simpler approaches such as the prominent \gls{pid} controller is inherently reactive such that an error has to occur before it can be remedied, a \gls{ml} based controller can act as a predictive system, a feature predominantly shared by approaches more computationally and knowledge intense such as the \gls{mpc}.
%\myworries{Referere til RL-bok her, f.eks. Sutton and Barto.}
\Gls{rl}~\cite{suttonbarto} is a subfield of \gls{ml} concerned  with  how  agents  should  act  in  order  to maximize some measure of utility, and how they can learn this behaviour from  interacting  with their environment. Control has historically been viewed as a difficult application of \gls{rl} due to the continuous nature of these problems' state and action spaces. Furthermore, the task has to be sufficiently nonlinear and complex for \gls{rl} to be an appropriate consideration over conventional control methods in the first place. To apply tabular methods one would have to discretize and thus suffer from the consequences of the curse of dimensionality from a discretization-resolution appropriate to achieve acceptable control. The alternative to tabular methods require function approximation, which has to be sophisticated enough to handle the dynamics of the task, while having a sufficiently stable and tractable training process to offer convergence. \Glspl{nn} are one of few models which satisfy these criteria: they can certainly be made expressively powerful enough for many tasks, but achieving a stable training phase can be a great challenge. Advances in computation capability and algorithmic progress in \gls{rl}, reducing the variance in parameter updates, have made \glspl{dnn} applicable to \gls{rl} algorithms, spawning the field of \gls{drl}. \glspl{dnn} in \gls{rl} algorithms provide end-to-end learning of appropriate representations and features for the task at hand, allowing algorithms to solve classes of problems previously deemed unfit for \gls{rl}. \Gls{drl} has been applied to complex control  tasks such as motion control of robots~\cite{Zhang} as well as other tasks where formalizing a strategy with other means is difficult, e.g. game playing~\cite{mnih_human-level_2015}.

A central challenge with \gls{rl} approaches to control is the low sample efficiency of these methods, meaning they need a large amount of data before they can become proficient. Allowing the algorithm full control to learn from its mistakes is often not a viable option due to operational constraints such as safety, and simulations are therefore usually the preferred option. The simulation is merely an approximation of the true environment. The model errors, i.e. the differences between the simulator and the real world, is called the reality gap. If the reality gap is small, then the low sample efficiency of these methods is not as paramount, and the agent might exhibit great skill the first time it is applied to the real world.

The current state-of-the-art \gls{rl} algorithms in continuous state and action spaces, notably \gls{ddpg}~\cite{lillicrap_continuous_2015}, \gls{trpo}~\cite{schulman_trust_2015}, \gls{ppo}~\cite{schulman_proximal_2017} and \gls{sac}~\cite{haarnoja_soft_2018}, are generally policy-gradient methods, where some parameterization of the policy is iteratively optimized through estimating the gradients. They are model-free, meaning they make no attempt at estimating the state-transition function. Thus they are very general and can be applied to many problems with little effort, at the cost of lower sample efficiency. %They can be on-policy or off-policy, where the former optimizes the same policy it explores and gathers data with, while the latter has a separate behaviour policy and can therefore be trained in an offline manner with older and previously gathered data, lowering the amount of data required. 
These methods generally follow the actor-critic architecture, wherein the actor module, i.e. the policy, chooses actions for the agent and the critic module evaluates how good these actions are, i.e. it estimates the expected long term reward, which reduces variance of the gradient estimates.

The premise of this research was to explore the application of \gls{rl} methods to low-level control of fixed-wing \glspl{uav}, in the hopes of producing a proof-of-concept \gls{rl} controller capable of stabilizing the attitude of the \acrshort{uav} to a given attitude reference. To this end, an OpenAI Gym environment~\cite{brockman_openai_2016} with a flight simulator tailored to the Skywalker X8 flying wing was implemented, in which the \gls{rl} controller is tasked with controlling the attitude (the roll and pitch angles) as well as the airspeed of the aircraft. Aerodynamic coefficients for the X8 are given in~\cite{Gryte}. The flight simulator was designed with the goal of being valid for a wide array of flight conditions, and therefore includes additional nonlinear effects in the aerodynamic model. The software has been made openly available \cite{pfly, pfly_env}. Key factors impacting the final performance of the controller as well as the rate of progression during training were identified. To the best of the authors' knowledge, this is the first reported work to use \gls{drl} for attitude control of fixed-wing \glspl{uav}.

The rest of the paper is organized as follows. First, previous applications of \gls{rl} algorithms to \glspl{uav} are presented in Section \ref{sec:related}, and the aerodynamic model of the Skywalker X8 fixed-wing \gls{uav} is then introduced in Section \ref{sec:model}. Section \ref{sec:approach} outlines the approach taken to develop the \gls{rl} controller, presenting the configuration of the \gls{rl} algorithm and the key design decisions taken, and finally how the controller is evaluated. In Section \ref{sec:results}, the training process and its major aspects are presented and discussed, and the controller is evaluated in light of the approach described in the preceding section. Finally, Section \ref{sec:conclusion} offers some final remarks and suggestions for further work.

%\myworries{Point out that the method is model-free and training is done using simulation and not real life flight.}

%This template, modified in MS Word 2003 and saved as ÒWord 97-2003 \& 6.0/95 Ð RTFÓ for the PC, provides authors with most of the formatting specifications needed for preparing electronic versions of their papers. All standard paper components have been specified for three reasons: (1) ease of use when formatting individual papers, (2) automatic compliance to electronic requirements that facilitate the concurrent or later production of electronic products, and (3) conformity of style throughout a conference proceedings. Margins, column widths, line spacing, and type styles are built-in; examples of the type styles are provided throughout this document and are identified in italic type, within parentheses, following the example. Some components, such as multi-leveled equations, graphics, and tables are not prescribed, although the various table text styles are provided. The formatter will need to create these components, incorporating the applicable criteria that follow.

\section{UAV MODEL} \label{sec:model}
Following~\cite{Beard}, the UAV is modeled as a rigid body of mass $m$ with inertia tensor $\boldsymbol{I}$ and a body frame $\{b\}$ rigidly attached to its center of mass, moving relative to a north-east-down (NED) frame assumed to be inertial $\{n\}$. To allow for arbitrary attitude maneuvers during simulation, the attitude is represented using unit quaternions $\boldsymbol{q} = [\eta\,\,\,\epsilon_1\,\,\,\epsilon_2\,\,\,\epsilon_3]^T$ where $\boldsymbol{q}^T \boldsymbol{q} = 1$. The time evolution of the position $\boldsymbol{p}=[x\,\,\,y\,\,\,z]^T$ and attitude $\boldsymbol{q}$ of the UAV is governed by the kinematic equations
\begin{align}
\dot{\boldsymbol{p}}  &= \boldsymbol{R}^n_b (\boldsymbol{q}) \boldsymbol{v} \\
\dot{\boldsymbol{q}}  &= \frac{1}2 \begin{bmatrix} 
    0 & - \boldsymbol{\omega}^T \\
    \boldsymbol{\omega} & - \boldsymbol{S}(\boldsymbol{\omega})
    \end{bmatrix} \boldsymbol{q} 
\end{align}
where $\boldsymbol{v} = [u\,\,\,v\,\,\,w]^T$ and $\boldsymbol{\omega} = [p\,\,\,q\,\,\,r]^T$ are the linear and angular velocities, respectively, and $\boldsymbol{S}(a)$ is the skew-symmetric matrix
\begin{equation}
    \boldsymbol{S}(a) = -\boldsymbol{S}^T(a) = \begin{bmatrix} 0 & -a_3 & a_2 \\
    a_3 & 0 & -a_1 \\
    -a_2 & a_1 & 0\end{bmatrix}
\end{equation}

The attitude can also be represented using Euler angles $\boldsymbol{\Theta} = [\phi\,\,\,\theta\,\,\,\psi]^T$, where $\phi$, $\theta$, $\psi$ are the roll, pitch and yaw angles respectively. Euler angles will be used for plotting purposes in later sections, and also as inputs to the controllers. Algorithms to convert between unit quaternions and Euler angles can be found in~\cite{Beard}.

The rotation matrix $\boldsymbol{R}^n_b$ transforms vectors from $\{b\}$ to $\{n\}$ and can be calculated from $\boldsymbol{q}$ using~\cite{Fossen2011}
\begin{equation}
\boldsymbol{R}^n_b(\boldsymbol{q}) = \boldsymbol{I}_{3 \times 3} + 2 \eta \boldsymbol{S}(\boldsymbol{\epsilon}) + 2 \boldsymbol{S}^2(\boldsymbol{\epsilon})
\end{equation}
where $\boldsymbol{I}_{3 \times 3}$ is the 3 by 3 identity matrix and $\boldsymbol{\epsilon} = [\epsilon_1 \,\, \epsilon_2 \,\, \epsilon_3]^T$.

The rates of change of the velocities $\boldsymbol{v}$ and $\boldsymbol{\omega}$ are given by the Newton-Euler equations of motion:
\begin{align}
m \dot{\boldsymbol{v}} + \boldsymbol{\omega} \times m\boldsymbol{v}&= \boldsymbol{R}_b^{n}(\boldsymbol{q})^Tm\boldsymbol{g}^n + \boldsymbol{F}_{prop}+\boldsymbol{F}_{aero}\\
\boldsymbol{I}\dot{\boldsymbol{\omega}} + \boldsymbol{\omega} \times \boldsymbol{I} \boldsymbol{\omega}&=  \boldsymbol{M}_{prop} + \boldsymbol{M}_{aero}
\end{align}
where $\boldsymbol{g}^n = [0\,\,\,0\,\,\,g]^T$ and $g$ is the acceleration of gravity. Apart from gravity, the UAV is affected by forces and moments due to aerodynamics and propulsion, which are explained in the next sections. All velocities, forces and moments are represented in the body frame.

%\todo[inline]{Står det noe om simulering av disse likningene? ode45, normalisering a q hver timestep etc.?}
\subsection{Aerodynamic Forces and Moments}
The UAV is flying in a wind field decomposed into a steady part $\boldsymbol{v}_{w_s}^n$ and a stochastic part $\boldsymbol{v}^b_{w_g}$ representing gusts and turbulence. The steady part is represented in $\{n\}$, while the stochastic part is represented in $\{b\}$. Similarly, rotational disturbances are modeled through the wind angular velocity $\boldsymbol{\omega}_{w}$. The relative (to the surrounding air mass) velocities of the UAV is then defined as:
\begin{align}
\boldsymbol{v}_r &= \boldsymbol{v} - \boldsymbol{R}^n_b(\boldsymbol{q})^T\boldsymbol{v}_{w_s} - \boldsymbol{v}_{w_g} = \begin{bmatrix}
u_r \\ v_r \\ w_r
\end{bmatrix} \\
\boldsymbol{\omega}_r &=
\boldsymbol{\omega}
-
\boldsymbol{\omega}_{w}
=
\begin{bmatrix}
p_r \\ q_r \\ r_r
\end{bmatrix}
\end{align}
From the relative velocity we can calculate the airspeed $V_a$, angle of attack $\alpha$ and sideslip angle $\beta$:
\begin{align}
V_a &= \sqrt{u_r^2 + v_r^2 + w_r^2 } \\
\alpha &= \tan^{-1}\left(\frac{u_r}{w_r}\right)  \\
\beta &= \sin^{-1}\left(\frac{v_r}{V_a}\right)
\end{align}
The stochastic components of the wind, given by $\boldsymbol{v}_{w_g}=[u_{w_g}\,\,\,v_{w_g}\,\,\,w_{w_g}]^T$ and $\boldsymbol{\omega}_{w}=[p_w\,\,\,q_w\,\,\,r_w]^T$ are generated by passing white noise through shaping filters given by the Dryden velocity spectra~\cite{dryden}\cite{dryden_matlab}.

The aerodynamic forces and moments are formulated in terms of aerodynamic coefficients $C_{(*)}$ that are, in general, nonlinear functions of $\alpha$, $\beta$ and $\boldsymbol{\omega}_r$, as well as control surface deflections. Aerodynamic coefficients are taken from~\cite{Gryte}, based on wind tunnel experiments of the Skywalker X8 flying wing as well as a Computational Fluid Dynamics (CFD) code. The X8 is equipped with right and left elevon control surfaces. Note that there is no tail or rudder. Even though the vehicle under consideration has elevons, in~\cite{Gryte} the model is parameterized in terms of "virtual" aileron and elevator deflections $\delta_a$ and $\delta_e$. These are related to elevon deflections through the transformation
\begin{equation}\label{eq:elevon_mapping}
\begin{bmatrix}
\delta_a \\ \delta_e
\end{bmatrix} = \begin{bmatrix}
-0.5 & 0.5 \\ 0.5 & 0.5
\end{bmatrix} \begin{bmatrix}
\delta_{e,r} \\ \delta_{e,l}
\end{bmatrix}
\end{equation}
where $\delta_{e,r}$ and $\delta_{e,l}$ are right and left elevon deflections, respectively.

The aerodynamic forces are described by
\begin{align}
\boldsymbol{F}_{aero} &= \boldsymbol{R}^b_w(\alpha,\beta)\begin{bmatrix} 
-D \\Y \\ -L
\end{bmatrix} \\
\begin{bmatrix} 
D  \\ Y \\ L
\end{bmatrix} &= \frac{1}{2}\rho V_a^2S  \begin{bmatrix} C_D(\alpha,\beta,q_r,\delta_{e}) \\ C_Y(\beta,p_r,r_r,\delta_{a}) \\ C_L(\alpha,q_r,\delta_{e})\end{bmatrix} \\
\boldsymbol{M}_{aero}  &= \frac{1}{2}\rho V_a^2S\begin{bmatrix}
bC_l(\beta,p_r,r_r,\delta_{a}) \\ cC_m(\alpha,q_r,\delta_{e}) \\ bC_n(\beta,p_r,r_r,\delta_{a})
\end{bmatrix}
\end{align}
where $\rho$ is the density of air, $S$ is the wing planform area, $c$ is the aerodynamic chord, and $b$ the wingspan of the UAV.
The rotation matrix transforming the drag force $D$, side force $Y$ and lift force $L$ from the wind frame to the body frame is given by:
\begin{equation}
\boldsymbol{R}^b_w(\alpha,\beta) = \left[\begin{smallmatrix}
\cos(\alpha)\cos(\beta) & \cos(\alpha)\sin(\beta) & -\sin(\alpha) \\-\sin(\beta) & \cos(\beta) & 0 \\
\cos(\beta)\sin(\alpha) & \sin(\alpha)\sin(\beta) & \cos(\alpha)
\end{smallmatrix}\right]
\end{equation}

The model in~\cite{Gryte} has similar structure to the linear coefficients in~\cite{Beard}, but has added quadratic terms in $\alpha$ and $\beta$ to the drag coefficient $C_D$. In addition, $C_D$ is quadratic in the elevator deflection $\delta_e$. In this paper, as an attempt to extend the range of validity of the model, the lift, drag and pitch moment coefficients in~\cite{Gryte} are extended using nonlinear Newtonian flat plate theory from \cite{Beard} and \cite{Gryte2015}. This makes the lift, drag and pitch coefficients nonlinear in angle of attack by blending between the linear models which are valid for small angles, and the flat plate models which are only valid for large angles. While the linear models are based on physical wind-tunnel experiments and CFD, the nonlinear models have not been validated experimentally.

\subsection{Propulsion Forces and Moments}
Assuming the propeller thrust is aligned with the x-axis of $\{b\}$, we can write
\begin{equation}
\boldsymbol{F}_{prop} = \begin{bmatrix} T_p \\ 0 \\ 0\end{bmatrix}
\end{equation}
The propeller thrust $T_p$ is given by~\cite{fitzpatrick} as presented in~\cite{beard2}:
\begin{align}
V_d   &= V_a + \delta_t (k_m - V_a) \label{eq:vd}\\
T_p   &= \frac{1}{2}\rho S_p C_p V_d \left(V_d - V_a \right) \label{eq:Tp}
\end{align}
where $V_d$ is the discharge velocity of air from the propeller, $k_m$ is a motor constant, $S_p$ is the propeller disc area, $C_p$ is an efficiency factor, and $\delta_t \in [0,1]$ is the throttle. The parameters in \eqref{eq:vd} and \eqref{eq:Tp} for a typical X8 motor/propeller configuration are given in~\cite{Coates}, which are based on wind tunnel experiments.

The propeller moments are given by
\begin{equation}
\boldsymbol{M}_{prop} = \begin{bmatrix} -k_Q (k_\Omega \delta_t)^2 \\ 0 \\ 0\end{bmatrix}
\end{equation}
where $k_\Omega=797.1268$ and $k_Q=1.1871\mathrm{e}{-6}$, which are based on the same experimental data used in~\cite{Coates}. Gyroscopic moments are assumed negligible.

\subsection{Actuator Dynamics and Constraints}
Denoting commands with superscript c, the elevon control surface dynamics are modeled by rate limited and saturated second-order integrators similar to \cite{prasad_pradeep}:
\begin{equation}
    \frac{\delta_{e,i}(s)}{\delta_{e,i}^c(s)} = \frac{\omega_0^2}{s^2 + 2\zeta \omega_0 s + \omega_0^2}
\end{equation}
for $i = r,l$, where $\omega_0 = 100$ and $\zeta = \frac{1}{\sqrt{2}}$. The angular deflections and rates are constrained to $\pm30$ degrees and $\pm200$ degrees per second, respectively.

The throttle dynamics are given by the first order transfer function~\cite{Gryte2015}
\begin{equation}
    \frac{\delta_t(s)}{\delta_t^c(s)} = \frac{1}{T s + 1}
\end{equation}
where $T = 0.2$.

\section{BACKGROUND} \label{sec:background}
\subsection{Reinforcement Learning}
\myworries{Korte ned og flytte til introduction?}
\gls{rl} is a subfield of \gls{ml} concerned with how agents should act in order to maximize some measure of utility, and how they can learn this behaviour from interacting with their environment. Unlike supervised learning, where the agent is supplied with the solution to every problem presented to it, the \gls{rl} agent receives only intermittent rewards, and it is the agents task to determine which actions are responsible for the reward, and subsequently determine a policy on how to act in the future. In this way, \gls{rl} is more powerful than supervised learning as the engineer does not need to have the solution to every problem in advance, but merely be able to formulate some measure of the utility of the agents current state. An example of such a situation is motion control of robots, where it is very difficult to formulate directions to every single step involved in maneuvering around, but it is comparatively easy to provide a utility measure in terms of forward progress towards some goal.

\gls{rl} is generally framed as a \gls{mdp}, defined by a set of states $S$, a set of actions $A(s)$ available in each state, a reward function $R(s)$, as well as a state transition function $P(a, s)$, giving the (probability of the) resulting state from choosing action $a$ in the current state $s$, with the goal being to find a policy $\pi$ which determines the optimal action to take in any given state. Typically, one also defines a value function $V(a, s) = E[\sum_{t=0}^\infty \gamma^t R_t]$, which describes the long term expected reward (possibly discounted through $\gamma \in (0, 1]$) of the action. This formulation has clear similarities with the typical control theory problem formulation, given by a systems states, the actions available to the controller through the systems actuators, the cost function and the differential equations governing the evolution of the systems state as influenced by the controllers actions.

Continuous control has historically been viewed as a difficult application of \gls{rl} due to methods, such as the popular Q-learning method, generally being table-based, and the resulting explosion of size-complexity following the curse of dimensionality as the state and action spaces are discretized to be made compatible. A better solution to the compatibility problem was found in using function approximators for the long-term reward of each state-action pair, instead of memorizing every pair as was the case before. \glspl{dnn} were found to be excellent function approximators due to being very expressively powerful and flexible, while only making a few assumptions about the function such as smoothness and hierarchical composition. The combination of \glspl{dnn} with \gls{rl} algorithms spawned the field of \gls{drl}, which have been successfully applied to many challenging problems previously deemed unsuitable for \gls{rl}. 

The current state-of-the-art \gls{rl} algorithms for continuous control, notably \gls{ddpg} \cite{lillicrap_continuous_2015}, \gls{trpo} \cite{schulman_trust_2015} and \gls{ppo} \cite{schulman_proximal_2017} are generally policy-gradient methods, where some parameterization of the policy is optimized directly as opposed to first estimating the values of each action-state pair and constructing the policy based on these estimates. They are model-free, meaning they make no attempt at estimating the state-transition function, thus they are very general and can be applied to many problems with little effort. On the other hand, this comes at the cost of low sample efficiency as each data point can only be used to update estimates of the policy, as opposed to providing data for each differential equation in its model. They can be either on-policy, i.e. the same policy that is used for choosing optimal actions is used to gather data and explore, or off-policy where some other policy with certain desirable characteristics can be used to gather data. Crucially, on-policy methods require fresh data samples that are relevant for its current behaviour to train its policy, while off-policy methods can store samples for later use, helping sample efficiency and reducing training time. These methods all follow the actor-critic architecture, wherein the actor module, that is the policy, chooses actions for the agent and the critic module evaluates how good these actions are, i.e. it estimates the expected long term reward. These two modules are typically realized as \glspl{nn} in \gls{drl} methods.

Since \gls{rl} methods are generally very sample inefficient, and real world data is often difficult to obtain, either because it is too time consuming or that the equipment is too valuable to risk damage with partly trained agents in control. Simulations are therefore an attractive alternative to provide a risk-free environment for the agent to train in. The simulated environment is however only an approximation of the environment that the agent is intended to operate in, an approximation that can be of varying accuracy. The magnitude of the discrepancy between the simulation and the real world is called the reality gap. The smaller the reality gap, the less sample complexity matters, as any amount of data desired is available through simulations at very low cost.

\myworries{write about exploration vs. exploitation?}

\subsection{The Proximal Policy Optimization Algorithm}
\myworries{Flytte til approach/method?}
\gls{ppo} is a model-free, on-policy, actor-critic, policy-gradient method notable for its high performance, low complexity, better sample complexity and low computational intensity compared to similar methods. It aims to retain the reliable performance of \gls{trpo} algorithms, which guarantee monotonic improvements by considering the KL divergence of new and old policy distribution, while only using first-order optimization. 

Policy gradient algorithms work by estimating the policy gradient, and then applying a gradient ascent algorithm to the gradient estimate. Two policy networks are maintained at all times: the current policy that we want to refine $\pi_{\theta}$, and the old policy which we use to explore and gather samples, $\pi_{\theta_{old}}$. Having two separate networks helps in sample efficiency. As the new behaviour policy is refined, the two policies will diverge, increasing variance of the estimation, and the old policy is therefore periodically updated to match the new policy. As in \gls{trpo}, \gls{ppo} optimizes a surrogate objective function, and defines the following quantities: the probability ratio $r_t(\theta) = \frac{\pi_\theta(a_t, s_t)}{\pi_{\theta_{old}}(a_t, s_t}$ which describes the changes in the agents action preferences under the new and old policy, as well as the advantage function $A = Q(s, a) - B(s)$ which measures how good an action is compared to the average of the actions available in that state. $B(s)$ is a baseline estimate of the long term rewards of the state, usually taken to be $B(s) = V(s)$, i.e. the value function of the state as estimated by the critic network. \gls{ppo} further proposes maximizing the objective function:

\begin{equation}
    L(\theta) = E_t[min(r_t(\theta)\Hat{A}_t, clip(r_t(\theta), 1 - \epsilon, 1 + \epsilon)\hat{A}_t)]
\end{equation}

%\myworries{Where the t denotes we sum over all T actions?}

This surrogate objective function limits the change in the policy by clipping the possible changes to the region $[1 - \epsilon, 1 + \epsilon]$ ($\epsilon$ is typically set to 0.2). Thus, the algorithm is not too greedy in favoring actions with a positive advantage function (i.e. actions better than its peers), and not too quick to avoid actions with a negative advantage function from a small set of samples. The minimum operator ensures that this objective function is a lower bound on the unclipped objective, and eliminates increased preference for actions with negative advantage function.
% TODO: maybe include last objective function
\myworries{Burde vi inkludere pseudocode av algoritmen?}

\section{RELATED WORK} \label{sec:related}
%Fixed-wing aircraft are normally underactuated vehicles, i.e. not actuated in all directions in 3D space. Guidance of such vehicles along a desired path is typically performed by controlling the attitude (orientation) and speed. For most aircraft, the magnitude of the lift force is typically larger than the thrust force (\myworries{kilde}). Heading changes are therefore most efficiently done by banking and thus tilting the lift force out of the vertical plane, creating a centripetal force for turning (bank-to-turn). Similarily, altitude changes can be made by controlling the flight path angle through the pitch angle. The inner-loop consists of linear \gls{pid} controllers for roll and pitch angles. The primary controls for roll are the aileron control surfaces, while an elevator provides pitching moments. Some \glspl{uav} also have a tail with a rudder that can produce yawing moments, which are typically used to minimize sideslip when turning (coordinated turn).
In general, the application of \gls{rl} to \gls{uav} platforms has been limited compared to other robotics applications, due to data collection with \gls{uav} systems carrying significant risk of fatal damage to the aircraft. \Gls{rl} have been proposed as a solution to many high level tasks for \glspl{uav} such as the higher level path planning and guidance tasks, alongside tried and tested traditional controllers providing low-level stabilization. In the work of~\citet{gandhiLearning2017} a \gls{uav} is trained to navigate in an indoor environment by gathering a sizable dataset consisting of crashes, giving the \gls{uav} ample experience of how NOT to fly. In~\cite{Han}, the authors tackle the data collection problem by constructing a pseudo flight environment in which a fixed-wing \gls{uav} and the surrounding area is fitted with magnets, allowing for adjustable magnetic forces and moments in each \gls{dof}. In this way, the \gls{uav} can be propped up as one would do when teaching a baby to walk, and thereby experiment without fear of crashing in a setting more realistic than simulations.

\citet{Imanberdiyev2016} developed a model-based \gls{rl} algorithm called TEXPLORE to efficiently plan trajectories in unknown environments subject to constraints such as battery life. In~\cite{ZhangMPC}, the authors use \pgls{mpc} to generate training data for \pgls{rl} controller, thereby guiding the policy search and avoiding the potentially catastrophic early phase before an effective policy is found. Their controller generalizes to avoid multiple obstacles, compared to the singular obstacle avoided by the \gls{mpc} in training, does not require full state information like the \gls{mpc} does, and is computed at a fraction of the time. With the advent of \gls{drl}, it has also been used for more advanced tasks such as enabling intelligent cooperation between multiple \glspl{uav}~\cite{Hung2017}, and for specific control problems such as landing~\cite{Polvara2018}. \Gls{rl} algorithms have also been proposed for attitude control of other autonomous vehicles, including satellites~\cite{Xu2018} and underwater vehicles.~\citet{Carlucho2018} applies an actor-critic \gls{drl} algorithm to low-level attitude control of an \gls{auv} --- similar to the proposed method in this paper --- and find that the derived control law transfers well from simulation to real world experiments.

Of work addressing problems more similar in nature to the one in this paper, i.e. low-level attitude control of \glspl{uav}, one can trace the application of \gls{rl} methods back to the works of~\citet{Bagnell} and~\citet{Ng}, both focusing on helicopter \glspl{uav}. Both employed methods based on offline learning from data gathered by an experienced pilot, as opposed to the online self-learning approach proposed in this paper. The former focuses on control of a subset of the controllable states while keeping the others fixed, whereas the latter work extends the control to all six degrees of freedom. In both cases, the controllers exhibit control performance exceeding that of the original pilot when tested on real \glspl{uav}. In~\cite{Abbeel}, the latter work was further extended to include difficult aerobatic maneuvers such as forward flips and sideways rolls, significantly improving upon the state-of-the-art.~\citet{Cory2008} presents experimental data of a fixed-wing \gls{uav} perching maneuver using an approximate optimal control solution. The control is calculated using a value iteration algorithm on a model obtained using nonlinear function approximators and unsteady system identification based on motion capture data. %The results demonstrate that the glider is able to exploit the increased pressure drag resulting from high angles of attack to achieve a high-speed perching maneuver. 
\citet{Bou-Ammar2010} compared \pgls{rl} algorithm using \gls{fvi} for approximation of the value function, to a non-linear controller based on feedback linearization, on their proficiency in stabilizing a quadcopter \gls{uav} after an input disturbance. They find the feedback-linearized controller to have superior performance. Recently,~\citet{koch_reinforcement_2018} applied three state-of-the-art \gls{rl} algorithms to control the angular rates of a quadcopter \gls{uav}. They found \gls{ppo} to perform the best of the \gls{rl} algorithms, outperforming the \gls{pid} controller on nearly every metric. 

%\myworries{Both experiment with the methods in practice and find that the algorithms are capable of control performance exceeding that of the original pilot.}
\section{METHOD} \label{sec:approach}
\gls{ppo} was the chosen \gls{rl} algorithm for the attitude controller for several reasons: first, \gls{ppo} was found to be the best performing algorithm for attitude control of quadcopters in~\cite{koch_reinforcement_2018}, and secondly, \gls{ppo}'s hyperparameters are robust for a large variety of tasks, and it has high performance and low computational complexity. It is therefore the default choice of algorithm in OpenAIs projects.

The objective is to control the \gls{uav}'s attitude, so a natural choice of controlled variables are the roll, pitch and yaw angles. However, the yaw angle of the aircraft is typically not controlled directly, but through the yaw-rate that depends on the roll angle. In addition, it is desirable to stay close to some nominal airspeed to ensure energy efficient flight, to avoid stall, and to maintain control surface effectiveness which is proportional to airspeed squared. The \gls{rl} controller is therefore tasked with controlling the roll and pitch angles, $\phi$ and $\theta$, and the airspeed $V_a$ to desired reference values. At each time step the controller receives an immediate reward, and it aims at developing a control law that maximizes the sum of future discounted rewards.

The action space of the controller is three dimensional, consisting of commanded virtual elevator and aileron angles as well as the throttle. Elevator and aileron commands are mapped to commanded elevon deflections using the inverse of the transformation given by~\eqref{eq:elevon_mapping}.

The observation vector (i.e. the input to the \gls{rl} algorithm) contains information obtained directly from state feedback of states typically measured by standard sensor suites. No sensor noise is added. To promote smooth actions it also includes a moving average of previous actuator setpoints. Moreover, since the policy network is a feed-forward network with no memory, the observation vector at each time step consists of these values for several previous time steps to facilitate learning of the dynamics.

%While the simulator has access to perfect estimates of all states in the simulation, feeding all this information to the controller would not be a realistic representation of real flight, and might not even be productive. The choice of which measurements go into the observation vector (the input to the \gls{rl} algorithm in the nomenclature of the OpenAI gym environment) is therefore based on what is reasonably attainable in the field, as well as the information's impact on controller performance.

\subsection{The Proximal Policy Optimization Algorithm}
\gls{ppo} is a model-free, on-policy, actor-critic, policy-gradient method. It aims to retain the reliable performance of \gls{trpo} algorithms, which guarantee monotonic improvements by considering the \gls{kl} divergence of policy updates, while only using first-order optimization. In this section, $\pi$ is the policy network (that is, the control law) which is optimized wrt. its parameterization $\theta$,\footnote{$\theta$ is used in this section as it is the established nomenclature in the machine learning field, but will in the rest of the article refer to the pitch angle.} in this case the \gls{nn} weights. The policy network takes the state, $s$, as its input, i.e. the observation vector, and outputs an action, $a$, consisting of the elevator, aileron and throttle setpoints. For continuous action spaces, the policy network is tasked with outputting the moments of a probability distribution, in this case the means and variances of a multivariate Gaussian, from which actions are drawn. During training, actions are randomly sampled from this distribution to increase exploration, while the mean is taken as the action when training is completed. 

Policy gradient algorithms work by estimating the policy gradient, and then applying a gradient ascent algorithm to the gradient estimate. The gradients are estimated in \pgls{mc} fashion by running the policy in the environment to obtain samples of the policy loss $J(\theta)$ and its gradient~\cite{suttonbarto}:\footnote{$\tau$ represents trajectories of the form $(s_1, a_1, s_2, a_2, \dots, s_T, a_T)$} 

\begin{align}
    J(\theta) &= \mathbb{E}_{\tau \sim \pi_{\theta}(\tau)} \left[\sum_t R(s_t, a_t)\right] \label{eq:policy_loss} = \mathbb{E}_{\tau \sim \pi_{\theta}(\tau)}[R(\tau)] \\
    \nabla_\theta J(\theta) &= \mathbb{E}_{\tau \sim \pi_{\theta}(\tau)} \left[\left(\sum_{t=1}^T \nabla_\theta \log \pi_\theta (a_{t}|s_{t})\right) R(\tau)\right] \label{eq:policy_grad}
\end{align}

In practice, these gradients are obtained with automatic differentiation software on a surrogate loss objective, whose gradients are the same as \eqref{eq:policy_grad}, and are then backpropagated through the \gls{nn} to update $\theta$. 

The central challenge in policy gradient methods lie in reducing the variance of the gradient estimates, such that consistent progress towards better policies can be made. The actor-critic architecture makes a significant impact in this regard, by reformulating the reward signals in terms of advantage:

\begin{align}
    Q^\pi(s, a) &= \sum_t \mathbb{E}_{\pi_{\theta}} [R(s_t, a_t) \,|\, s, a] \\
    V^\pi(s) &= \sum_t \mathbb{E}_{\pi_{\theta}} [R(s_t, a_t) \, | \, s] \\
    A^\pi(s, a) &= Q^\pi(s, a) - V^\pi(s) \label{eq:advantage}
\end{align}

The advantage function \eqref{eq:advantage} measures how good an action is compared to the other actions available in the state, such that good actions have positive rewards, and bad actions have negative rewards. One thus has to be able to estimate the average reward of the state, i.e. the value function $V(s)$.\footnote{The value function $V^{\pi}(s)$ is the expected long term reward of being in state $s$ and then following policy $\pi$, as opposed to the $Q^\pi(s, a)$-function which focuses on the long term reward of taking a specific action in the state, and then following the policy.} This is the job of the critic network, a separate \gls{nn} trained in a supervised manner to predict the value function with ground truth from the reward values in the gathered samples. Several improvements such as \gls{gae} are further employed to reduce variance of the advantage estimates. \gls{ppo} also makes use of several actors simultaneously gathering samples with the policy, to increase the sample batch size.

\gls{ppo} maximizes the surrogate objective function 

\begin{equation}
    L(\theta) = \hat{\mathbb{E}}_t\left[\min\left(r_t(\theta)\Hat{A}_t, \clip\left(r_t(\theta), 1 - \epsilon, 1 + \epsilon\right)\hat{A}_t\right)\right] \label{eq:l_theta}
\end{equation}

in which $\hat{A}$ and $\hat{\mathbb{E}}$ denotes the empirically obtained estimates of the advantage function and expectation, respectively, and $r_t(\theta)$ is the probability ratio

\begin{equation}
     r_t(\theta) = \frac{\pi_\theta(a_t, s_t)}{\pi_{\theta_{old}}(a_t, s_t)} \label{eq:prob_ratio}
\end{equation}

Vanilla policy gradients require samples from the policy being optimized, which after a single optimization step are no longer usable for the improved policy. For increased sample efficiency, \gls{ppo} uses importance sampling to obtain the expectation of samples gathered from an old policy $\pi_{\theta_{old}}$ under the new policy we want to refine $\pi_{\theta}$. In this way, each sample can be used for several gradient ascent steps. As the new policy is refined, the two policies will diverge, increasing variance of the estimation, and the old policy is therefore periodically updated to match the new policy. For this approach to be valid, the state transition function must be similar between the two policies, which is ensured by clipping the probability ratio \eqref{eq:prob_ratio} to the region $[1 - \epsilon,\enspace1 + \epsilon]$.\footnote{The clip operator saturates the variable in the first argument between the values supplied by the two following arguments.} This also gives a first-order approach to trust region optimization: The algorithm is not too greedy in favoring actions with positive advantage, and not too quick to avoid actions with a negative advantage function from a small set of samples. The minimum operator ensures that the surrogate objective function is a lower bound on the unclipped objective, and eliminates increased preference for actions with negative advantage function. \Gls{ppo} is outlined in Algorithm \ref{alg:ppo}.

%\myworries{Where the t denotes we sum over all T actions?}
% TODO: maybe include last objective function

\begin{algorithm}
    \SetAlgoLined
    \For{iteration=1, 2, \dots 
    }{
    \For{actor=1, 2, \dots, N}{
    Run policy $\pi_{\theta_{old}}$ in environment for T time steps \\
    Compute advantage estimates $\hat{A}_t$ for $t=1, 2, \dots, T$
    }
    Optimize surrogate L wrt. $\theta$. \\
    $\theta_{old} \gets \theta$
    }
    \caption{PPO}
    \label{alg:ppo}
\end{algorithm}

\subsection{Action Space}\label{sec:action_space}
% TODO: add action u delta cost
A known issue in optimal control is that while continually switching between maximum and minimum input is often optimal in the sense of maximizing the objective function, in practice it wears unnecessarily on the actuators. Since \gls{ppo} during training samples its outputs from a Gaussian distribution, a high variance will generate highly fluctuating actions. This is not much of a problem in a simulator environment but could be an issue if trained online on a real aircraft. \gls{ppo}'s hyperparameters are tuned wrt. a symmetric action space with a small range (e.g. -1 to 1). Adhering to this design also has the benefit of increased generality, training the controller to output actions as a fraction of maximal and minimal setpoints. The actions produced by the controller are therefore clipped to this range, and subsequently scaled to fit the actuator ranges as described in Section \ref{sec:model}.

\subsection{Training of Controller}

\begin{table}
    \vspace*{0.13cm} 
    \caption{Constraints and ranges for initial conditions and target setpoints used during training of controller.}
    \centering
    \ra{1.2}
    \begin{tabular}{@{}lrrr@{}} \toprule
    Variable & Initial Condition & Target \\  \midrule
    $\phi$ & $\pm 150\degree$ & $\pm 60\degree$ \\
    $\theta$ & $\pm 45\degree$ & $\pm 30\degree$ \\
    $\psi$ & $\pm 60\degree$ & - \\
    $\omega$ & $\pm 60 \thinspace \degree/\textrm{s}$ & - \\
    $\alpha$ & $\pm 26 \degree$ & - \\
    $\beta$ & $\pm 26 \degree$ & - \\
    $V_a$ & $12-30 \thinspace \textrm{m/s}$ & $12-30 \thinspace \textrm{m/s}$ \\
    %Wind & $[\pm 8, \pm 8, \pm 8] m/s$ & - & - \\
    %$\alpha$ & - & - & $\pm 85 \degree$ \\
    \bottomrule
    \end{tabular}
    \label{tab:training_variables}
\end{table}

The \gls{ppo} \gls{rl} controller was initialized with the default hyperparameters in the OpenAI Baselines implementation~\cite{stable-baselines}, and ran with 6 parallel actors. The controller policy is an extended version of the default two hidden layer, 64 nodes \gls{mlp} policy: The observation vector is first processed in a convolutional layer with three filters spanning the time dimension for each component, before being fed to the default policy. This allows the policy to construct functions on the time evolution of the observation vector, while scaling more favorably in parameter count with increasing observation vector size compared to a fully connected input layer.

The controller is trained in an episodic manner to assume control of an aircraft in motion and orient it towards some new reference attitude. Although the task at hand is not truly episodic in the sense of having natural terminal states, episodic training allows one to adjust episode conditions to suit the agents proficiency, and also admits greater control of the agents exploration of the state space. The initial state and reference setpoints for the aircraft are randomized in the ranges shown in Table \ref{tab:training_variables}. Episode conditions are progressively made more difficult as the controller improves, beginning close to target setpoints and in stable conditions, until finally spanning the entirety of Table \ref{tab:training_variables}. The chosen ranges allow the \gls{rl} controller to demonstrate that it is capable of attitude control, and facilitates comparison with the \gls{pid} controller as it is expected to perform well in this region. According to \cite{Beard}, a typical sampling frequency for autopilots is 100 Hertz, and the simulator therefore advances 0.01 seconds at each time step. Each episode is terminated after a maximum of 2000 time steps, corresponding to 20 seconds of flight time. No wind or turbulence forces are enabled during training of the controller.

% episodic training is good
% increasing difficulty
% goal termination

% TODO: i think we should remove this
%In order to encourage desirable behaviour and for stability of the simulator, some states were constrained. In particular, the sideslip angle $\beta$ was constrained to $\pm 85 \degree$, and the angular rates $\omega_{\{1, 2, 3\}}$ were constrained to $\pm \pi\thinspace rad/s$. The motivation behind constraining the angular rates is twofold, first the aerodynamic coefficients are only valid for small values of the angular rates, and second it helps guide the \gls{rl} controller towards reasonable behaviour.
% TODO: maybe address that negative rewards encourage reaching terminal fastest, but i dont have terminal state, so -\_('_')_/-
In accordance with traditional control theory, where one usually considers cost to be minimized rather than rewards to be maximized, the immediate reward returns to the \gls{rl} controller are all negative rewards in the normalized range of -1 to 0:

\begin{align}\label{eq:rew}
    R_{\phi} &= \clip\left(\frac{|\phi - \phi^{d}|}{\zeta_1}, 0, \gamma_1\right) \nonumber \\
    R_{\theta} &= \clip\left(\frac{|(\theta - \theta^{d})|}{\zeta_2}, 0, \gamma_2\right) \nonumber \\
    R_{V_a} &= \clip\left(\frac{|V_a - V_{a}^d|}{\zeta_3}, 0, \gamma_3\right) \nonumber \\
    R_{\delta^c} &= \clip\left(\frac{\sum_{j \in [a, e, t]}\sum_{i=0}^4 |\delta^c_{j_{t-i}} - \delta^c_{j_{t-1-i}}|}{\zeta_4}, 0, \gamma_4\right) \nonumber \\
    R_t &= -(R_{\phi} + R_\theta + R_{V_a} + R_{\delta^c})
\end{align}
\begin{align}
    \zeta_1 &= 3.3,~\zeta_2 = 2.25,~\zeta_3 = 25,~\zeta_4 = 60 \nonumber \\
    \gamma_1 &= 0.3, ~\gamma_2 = 0.3, ~~\gamma_3 = 0.3,\gamma_4 = 0.1 \nonumber
\end{align}

%\delta_{t-i}^c - \delta_{t-1-i}^c

In this reward function, $L_1$ was chosen as the distance metric between the current state and the desired state (denoted with superscript d).\footnote{The $L_1$ distance has the advantage of punishing small errors harsher than the $L_2$ distance, and therefore encourages eliminating small steady-state errors.} Furthermore, a cost is attached to changing the actuator setpoints to address oscillatory control behaviour. Commanded control setpoint of actuator $j$ at time step $t$ is denoted $\delta^c_{j_{t}}$, where $j \in [a,e,t]$. The importance of each component of the reward function is weighted through the $\gamma$ factors. To balance the disparate scales of the different components, the values are divided by the variables approximate dynamic range, represented by the $\zeta$ factors.

%Another important aspect of the reward structure is how to handle premature termination of the episode due to constraint violations. With no additional punishment for these situations, an alluring solution for the \gls{rl} controller would be to steer the aircraft towards one of the constraints in as few time steps as possible, as this would give a low total cost. To discourage these types of strategies the \gls{rl} controller receives an additional punishment equal to the amount of steps left in the simulation when the episode is terminated due to constraint violation, i.e. it receives the maximum punishment for each remaining time step. This solution breaks the paradigm of normalized rewards in the range of -1 to 0, but works well as the policy will receive very steep negative gradients for actions which lead to these constraint-violating trajectories, and thereby discourage pursuing them.

The components of the observation vector are expressed in different units and also have differing dynamic ranges. \Glspl{nn} are known to converge faster when the input features share a common scale, such that the network does not need to learn this scaling itself. The observation vector should therefore be normalized. This is accomplished with the VecNormalize class of~\cite{stable-baselines}, which estimates a running mean and variance of each observation component and normalizes based on these estimates.

\subsection{Evaluation}
Representing the state-of-the-art in model free control, fixed-gain \gls{pid} controllers for roll, pitch and airspeed were implemented to provide a baseline comparison for the \gls{rl} controller:
\begin{align}
\delta_t^c &=  - k_{p_V}(V_a - V_a^d) - k_{i_V}\int_{0}^{t}(V_a - V_a^d)d\tau \label{eq:PID_Va}\\
\delta_a^c &=  - k_{p_\phi}(\phi - \phi^d) - k_{i_\phi}\int_{0}^{t}(\phi - \phi^d)d\tau - k_{d_\phi}p \label{eq:PID_roll} \\
\delta_e^c &=  - k_{p_\theta}(\theta - \theta^d) - k_{i_\theta}\int_{0}^{t}(\theta - \theta^d)d\tau - k_{d_\theta}q \label{eq:PID_pitch}
\end{align}
The throttle is used to control airspeed, while virtual aileron and elevator commands are calculated to control roll and pitch, respectively.
The \gls{pid} controllers were manually tuned using a trial-and-error approach until achieving acceptable transient responses and low steady-state errors for a range of initial conditions and setpoints. Wind was turned off in the simulator during tuning. The integral terms in \eqref{eq:PID_Va}-\eqref{eq:PID_pitch} are implemented numerically using forward Euler. Controller gains are given in Table~\ref{tab:gains}. 

\begin{table} [htbp]
    \caption{PID controller parameters.}
    \ra{1.2}
	\centering
	%{@{} *1l| *2r @{}}
	\begin{tabular}{@{} *1l *1r *1l *1r @{}} \toprule
		%\begin{tabular}{@{}| *1l| *1l |@{}} \toprule
		Parameter  & Value & Parameter  & Value   \\ \midrule
		$k_{p_V}$	   &  \num{0.5}    & $k_{d_\phi}$   & \num{0.5}    \\ 
		$k_{i_V}$	   &  \num{0.1}    & $k_{p_\theta}$ & \num{-4}   \\
		$k_{p_\phi}$   &  \num{1}      & $k_{i_\theta}$ & \num{-0.75} \\	
		$k_{i_\phi}$   &  \num{0}      & $k_{d_\theta}$ & \num{-0.1} \\
		\bottomrule
	\end{tabular}
	\label{tab:gains} 
\end{table}

The same aerodynamic model that is used for training is also used for evaluation purposes, with the addition of disturbances in the form of wind to test generalization capabilities. The controllers are compared in four distinct wind and turbulence regions: light, moderate, severe and no turbulence. Each setting consists of a steady wind component, with randomized orientation and a magnitude of 7 m/s, 15 m/s, 23 m/s and 0 m/s respectively, and additive turbulence given by the Dryden turbulence model~\cite{dryden}. Note that a wind speed of 23 m/s is a substantial disturbance, especially when considering the Skywalker X8's nominal airspeed of 18 m/s. For each wind setting, 100 sets of initial conditions and target setpoints are generated, spanning the ranges shown in Table \ref{tab:training_variables}. The reference setpoints are set to 20-30 degrees and 3-4 m/s deviation from the initial state for the angle variables and airspeed, respectively. Each evaluation scenario is run for a maximum of 1500 time steps, corresponding to 15 seconds of flight time, which should be sufficient time to allow the controller to regulate to the setpoint.

The reward function is not merely measuring the proficiency of the \gls{rl} controller, but is also designed to facilitate learning. To compare, rank and evaluate different controllers, one needs to define additional evaluation criteria. To this end, the controllers are evaluated on the following criteria: \textbf{Success/failure}, whether the controller is successful in controlling the state to within some bound of the setpoint. The state must remain within the bounds for at least 100 consecutive time steps to be counted as a success. The bound was chosen to be $\pm 5\degree$ for the roll and pitch angles, and $\pm 2 \textrm{m/s}$ for the airspeed. \textbf{Rise time}, the time it takes the controller to reduce the initial error from 90 \% to 10 \%. As these control scenarios are not just simple step responses and may cross these thresholds several times during the episode, the rise time is calculated from the first time it crosses the lower threshold until the first time it reaches the upper threshold. \textbf{Settling time}, the time it takes the controller to settle within the success setpoint bounds, and never leave this bound again. \textbf{Overshoot}, the peak value reached on the opposing side of the setpoint wrt. the initial error, expressed as a percentage of the initial error. \textbf{Control variation}, the average change in actuator commands per second, where the average is taken over time steps and actuators. Rise time, settling time, overshoot and control variation are only measured when the episode is counted as a success. When comparing controllers, the success criterion is the most important, as it is indicative of stability as well as achieving the control objective. Secondly, low control variation is important to avoid unnecessary wear and tear on the actuators. While success or failure is a binary variable, rise time, settling time and overshoot give additional quantitative information on the average performance of the successful scenarios.

%most weight is attributed to the success criteria. Settling time measures how quickly the controller is able to stabilize at the target setpoints, and as such is rated as the second most important measure. A low control variation is indicative of a smooth trajectory toward the target, and thus more likely to transfer well to a real flight environment. Finally, rise time and overshoot describe similar aspects, indicating how aggressive the controller is, and as such are weighted the least.
\section{RESULTS AND DISCUSSION} \label{sec:results}
The controller was trained on a desktop computer with an i7-9700k CPU and an RTX 2070 GPU. The model converges after around two million time steps of training, which on this hardware takes about an hour. This is relatively little compared to other applications of \gls{drl}, and suggests that the \gls{rl} controller has additional capacity to master more difficult tasks. Inference with the trained model takes $800$ microseconds on this hardware, meaning that the \gls{rl} controller could reasonably be expected to be able to operate at the assumed autopilot sampling frequency of $100$ Hertz in flight.

\subsection{Key Factors Impacting Training}
The choice of observation vector supplied to the \gls{rl} controller proved to be significant for its rate of improvement during training and its final performance. It was found that reducing the observation vector to only the essential components, i.e. the current airspeed and roll and pitch angles, the current angular velocities of the \gls{uav}, and the state errors, helped the \gls{rl} controller improve significantly faster than other, larger observation vectors.\footnote{Essential here referring to the factors' impact on performance for this specific control task. One would for instance expect $\alpha$ and $\beta$ to be essential factors when operating in the more extreme and nonlinear regions of the state space.} Including values for several previous time steps (five was found to a good choice) further accelerated training, as this makes learning the dynamics easier for the memoryless feed-forward policy.

The reward function is one of the major ways the designer can influence and direct the behaviour of the agent. One of the more popular alternatives to $L_1$ norm and clipping to achieve saturated rewards are the class of exponential reward functions, and notably the Gaussian reward function as in \cite{Carlucho2018}. Analyzing different choices of the reward function was not given much focus as the original choice gave satisfying results.
\begin{table*}
    \vspace*{0.13cm} 
    \centering
    \caption{Performance metrics for the RL controller and the baseline PID controller on the evaluation scenarios. Both controllers exhibit strengths in different aspects --- the best value in each circumstance is shown in bold.}
    \ra{1.2}
    \begin{tabular}{@{}llrrrrrrrrrrrrrc@{}} \toprule
    & & \multicolumn{4}{c}{Success (\%)} & \multicolumn{3}{c}{Rise time (s)} & \multicolumn{3}{c}{Settling time (s)} & \multicolumn{3}{c}{Overshoot (\%)} & \multirow{2}{*}{\parbox{1.3cm}{\centering Control variation ($s^{-1}$)}} \\
    \cmidrule(lr){3-6}\cmidrule(lr){7-9}\cmidrule(lr){10-12}\cmidrule(l){13-15}
    Setting & Controller & $\phi$ & $\theta$ & $V_a$ & All & $\phi$ & $\theta$ & $V_a$ & $\phi$ & $\theta$ & $V_a$ & $\phi$ & $\theta$ & $V_a$ \\ \midrule
    \multirow{2}{*}{No turbulence} & RL & 100 & 100 & \textbf{100} & \textbf{100} & \textbf{0.265} & 0.661 & \textbf{0.825} & \textbf{1.584} & 1.663 & 2.798 & 21 & 24 & \textbf{31} & 0.517 \\
    & PID & 100 & 100 & 98 & 98 & 1.344 & \textbf{0.228} & 0.962 & 2.050 & \textbf{1.364} & \textbf{2.198} & \textbf{4} & \textbf{17} & 35 & \textbf{0.199} \\ 
    \midrule
    \multirow{2}{*}{Light turbulence} & RL & 100 & 100 & \textbf{100} & \textbf{100} & \textbf{0.210} & 0.773 & \textbf{0.744} & \textbf{1.676} & 1.806 & 2.738 & 28 & 33 & \textbf{36} & 0.748 \\
    & PID & 100 & 100 & 99 & 99 & 1.081 & \textbf{0.294} & 0.863 & 2.057 & \textbf{1.638} & \textbf{2.369} & \textbf{6} & \textbf{20} & 43 & \textbf{0.457} \\
    \midrule
    \multirow{2}{*}{Moderate turbulence} & RL & \textbf{100} & \textbf{100} & \textbf{98} & \textbf{98} & \textbf{0.192} & 1.474 & 0.934 & \textbf{2.167} & \textbf{2.438} & 4.085 & 54 & 54 & 74 & 0.913 \\
    & PID & 100 & 93 & 90 & 87 & 0.793 & \textbf{0.525} & \textbf{0.864} & 2.764 & 2.563 & \textbf{3.460} & \textbf{34} & \textbf{35} & \textbf{70} & \textbf{0.781} \\
    \midrule
    \multirow{2}{*}{Severe turbulence} & RL & \textbf{100} & \textbf{100} & \textbf{92} & \textbf{92} & \textbf{0.166} & 1.792 & 1.585 & \textbf{2.903} & \textbf{3.280} & 7.049 & 122 & 93 & 156 & \textbf{1.083} \\
    & PID & 99 & 96 & 87 & 86 & 0.630 & \textbf{0.945} & \textbf{1.343} & 3.576 & 5.256 & \textbf{5.470} & \textbf{92} & \textbf{80} & \textbf{122} & 1.117 \\
    \bottomrule
    \end{tabular}
    \label{tab:result}
\end{table*}

\subsection{Evaluation of Controller}
\begin{figure}[thpb]
  %\hspace*{-0.2cm} 
  \centering
  \includegraphics[scale=0.4]{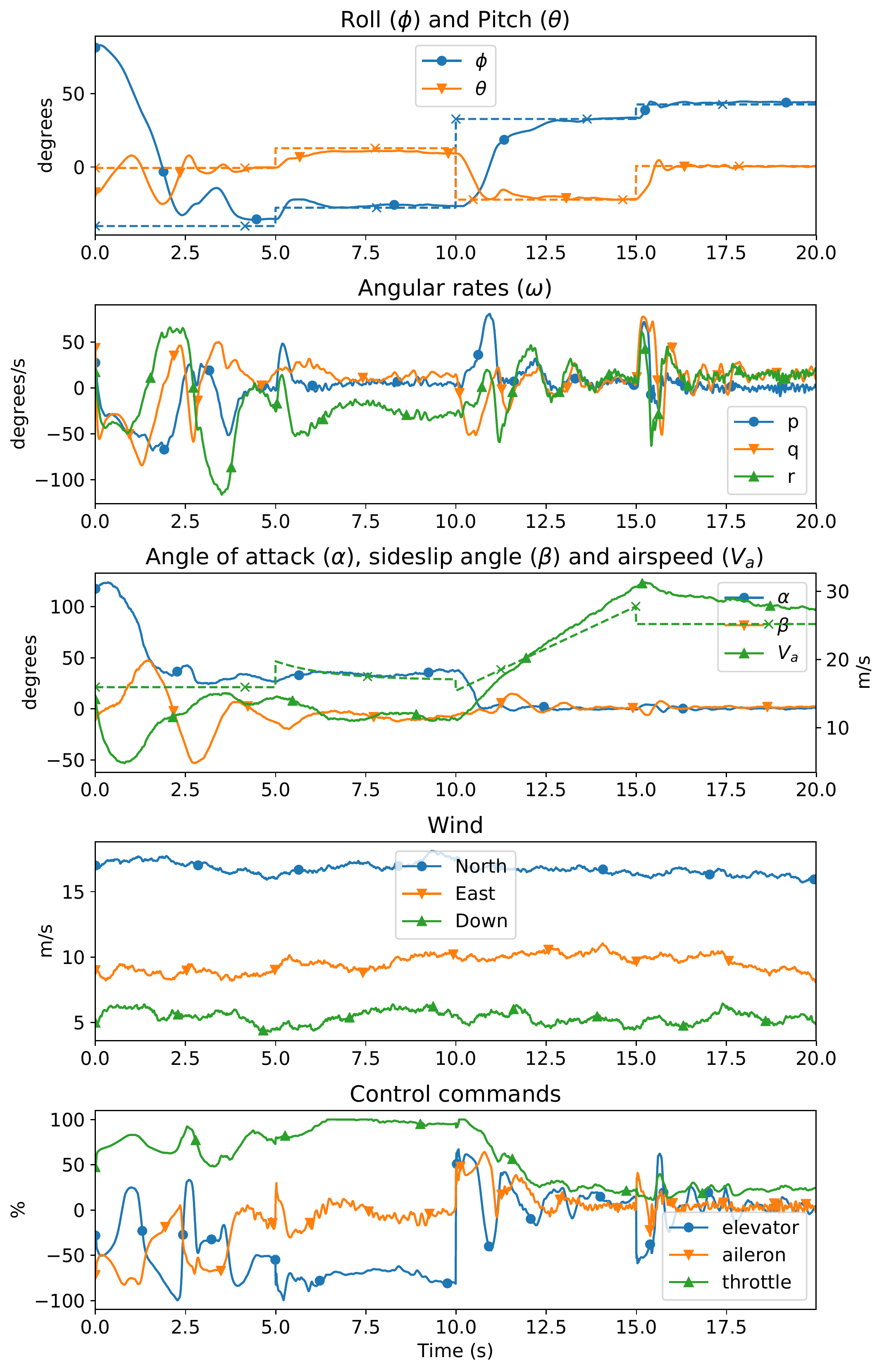}
  \caption{The \gls{rl} controller trained episodically with a single setpoint and no wind or turbulence generalizes well to many wind conditions and continuous tracking of setpoints (shown with dashed lines marked by crosses). Here subjected to severe wind and turbulence disturbances with a magnitude of 20 m/s.}
  \label{fig:continuous_task}
\end{figure}

\begin{figure}[thpb]
    \centering
    \includegraphics[scale=0.38]{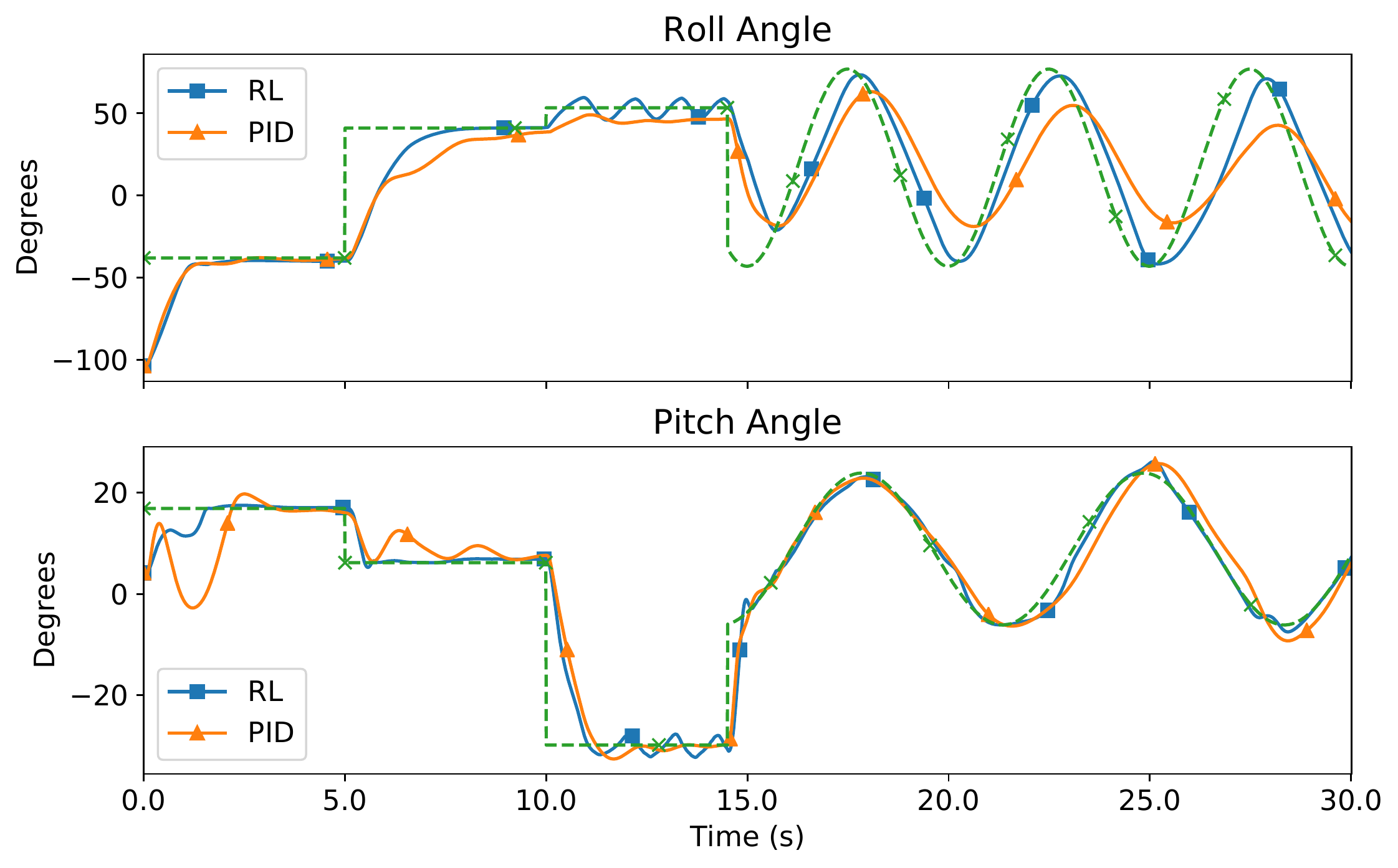}
    \caption{Comparison of the \gls{pid} and \gls{rl} controllers tasked with tracking the dashed green line.}
    \label{fig:comp}
\end{figure}

The \gls{rl} controller generalizes well to situations and tasks not encountered during training. Even though the controller is trained with a single setpoint for each episode, Figure \ref{fig:continuous_task} shows that the controller is perfectly capable of adapting to new setpoints during flight. This result was also found by \citet{koch_reinforcement_2018} for quadcopters. The generalization capability also holds true for unmodeled wind and turbulence forces. The controller is trained with no wind estimates present in the observation vector, and no wind forces enabled in the simulator, but as Table \ref{tab:result} shows it is still able to achieve tracking to the setpoint when steady wind and turbulence is enabled in the test environment. Table \ref{tab:result} should be read as a quantitative analysis of performance in conditions similar to normal operating conditions, while Figure \ref{fig:continuous_task} and \ref{fig:comp} qualitatively shows the capabilities of the controllers on significantly more challenging tasks.
%The unmodeled turbulence forces do however negatively impact the smoothness of the actions produced by the \gls{rl} controller. It is rapidly adjusting its control surfaces, most notably the aileron, to adjust for the unexpected deviations in the observed states --- an effect not exhibited when wind is not enabled. This could have adverse affects as described in Section \ref{sec:action_space}, and should be addressed further, for instance by additional cost on changes in actuator setpoints.

% RL controller is more successful, especially angle and pitch control. They are about equal in regulating speed, each having the edge in different states. PID has less overshoot and less command variation, which might transfer better to real flight. The command variation grows about equally with more disturbance. This test is on the PIDs terms. Neither has clear advantage over the other, but this is encouraging as this test is on the PIDs terms, and is not the region the RL is expected to make the greatest contribution. 

Table \ref{tab:result} shows that both controllers are generally capable of achieving convergence to the target for the evaluation tasks, with neither controller clearly outperforming the other. The \gls{rl} controller has an advantage over the \gls{pid} controller on the success criterion, and seems to be more robust to the turbulence disturbance. It is able to achieve convergence in the attitude states in all situations, unlike the \gls{pid} controller, and is also notably more successful in moderate and severe turbulence conditions. The \gls{pid} controller has considerably lower control variation for the simple settings with little or no wind, but its control variation grows fast with increasing disturbance. At severe turbulence the \gls{rl} controller has the least control variation.

The two controllers perform similarly wrt. settling time and rise time, each having the edge in different states under various conditions, while the \gls{pid} controller performs favorably when measured on overshoot. All in all, this is an encouraging result for the \gls{rl} controller, as it is able to perform similarly as the established \gls{pid} controller in its preferred domain, while the \gls{rl} controller is expected to make its greatest contribution in the more nonlinear regions of the state space. 

%Much the same can be concluded by inspecting Figure \ref{fig:comp}, which shows a comparison of the tracking performance of the two controllers. The \gls{rl} controller is faster at responding to changes in the reference --- indicative of its lower rise and settling times.

A comparison of the two controllers is shown in Figure \ref{fig:comp} on a scenario involving fairly aggressive maneuvers, which both are able to execute. Figure \ref{fig:continuous_task} and \ref{fig:comp} illustrate an interesting result, the \gls{rl} controller is able to eliminate steady state errors. While the \gls{pid} controller has integral action to mitigate steady-state errors, the control law of the \gls{rl} controller is only a function of the last few states and references. This might suggest that the \gls{rl} controller has learned some feed-forward action, including nominal inputs in each equilibrium state, thus removing steady-state errors in most cases. Another possibility is that steady-state errors are greatly reduced through the use of high-gain feedback, but the low control variation shown for severe turbulence in Table~\ref{tab:result} indicates that the gain is not excessive. Future work should include integral error states in the observations and evaluate the implications on training and flight performance.

\addtolength{\textheight}{-12cm}   % This command serves to balance the column lengths

%%%%%%%%%%%%%%%%%%%%%%%%%%%%%%%%%%%%%%%%%%%%%%%%%%%%%%%%%%%%%%%%%%%%%%%%%%%%%%%%

%%%%%%%%%%%%%%%%%%%%%%%%%%%%%%%%%%%%%%%%%%%%%%%%%%%%%%%%%%%%%%%%%%%%%%%%%%%%%%%%

%%%%%%%%%%%%%%%%%%%%%%%%%%%%%%%%%%%%%%%%%%%%%%%%%%%%%%%%%%%%%%%%%%%%%%%%%%%%%%%%
%\section*{APPENDIX}


%\section*{ACKNOWLEDGMENT}

%The preferred spelling of the word ÒacknowledgmentÓ in America is without an ÒeÓ after the ÒgÓ. Avoid the stilted expression, ÒOne of us (R. B. G.) thanks . . .Ó  Instead, try ÒR. B. G. thanksÓ. Put sponsor acknowledgments in the unnumbered footnote on the first page.

%%%%%%%%%%%%%%%%%%%%%%%%%%%%%%%%%%%%%%%%%%%%%%%%%%%%%%%%%%%%%%%%%%%%%%%%%%%%%%%%

%References are important to the reader; therefore, each citation must be complete and correct. If at all possible, references should be commonly available publications.
\bibliographystyle{unsrtnat}
\bibliography{references}{}

\end{document}